\documentclass{article}

\usepackage{microtype}
\usepackage{graphicx}
\usepackage{subfigure}
\usepackage{booktabs} 
\usepackage{caption}
\usepackage{amsmath}
\usepackage{amssymb}

\usepackage{hyperref}

\newcommand{\figleft}{{\em (Left)}}

\newcommand{\figright}{{\em (Right)}}
\newcommand{\figtop}{{\em (Top)}}
\newcommand{\figbottom}{{\em (Bottom)}}

\newtheorem{definition}{Definition}

\newtheorem{proposition}{Proposition}

\usepackage[accepted]{hill2019}
\icmltitlerunning{Knowledge-augmented Column Networks: Guiding Deep Learning with Advice}

\begin{document}

\twocolumn[
\icmltitle{Knowledge-augmented Column Networks: Guiding Deep Learning with Advice}



\icmlsetsymbol{equal}{*}

\begin{icmlauthorlist}
\icmlauthor{Mayukh Das}{to}
\icmlauthor{Devendra Singh Dhami}{to}
\icmlauthor{Yang Yu}{to}
\icmlauthor{Gautam Kunapuli}{to}
\icmlauthor{Sriraam Natarajan}{to}
\end{icmlauthorlist}

\icmlaffiliation{to}{University of Texas, Dallas}

\icmlcorrespondingauthor{Mayukh Das}{mayukh.das1@utdallas.edu}


\vskip 0.3in
]



\printAffiliationsAndNotice{}  

\begin{abstract}
Recently, deep models have had considerable success in several tasks, especially with low-level representations. 
However, effective learning from sparse noisy samples is a major challenge in most deep models, especially in domains with structured representations. Inspired by the proven success of human guided machine learning, we propose Knowledge-augmented Column Networks, a relational deep learning framework that leverages human advice/knowledge to learn better models in presence of sparsity and systematic noise. 

\end{abstract}

\section{Introduction}
The re-emergence of Deep Learning~\cite{DeepLearningBook2016} has demonstrated significant success in difficult real-world domains such as image \cite{krizhevsky2012imagenet}, audio \cite{audio} and video processing \cite{videoCVPR}. 
Deep Learning is recently being increasingly applied to structured domains, where the data is represented using {\em richer symbolic or graph features} to capture relational structure between entities and attributes in the domain. 
Such models are able to capture increasingly complex interactions between features with deeper layers. However, the combinatorial complexity of reasoning over a large number of relations and objects has remains a significant bottleneck to overcome. 

While recent work in relational deep learning seeks to 
address the problem of faithful modeling of relational structure
\cite{KazemiPoole18-RelNNs,SourekEtAl-15-LRNNs,KaurEtAl18-RRBM},  
we focus on {\bf Column Networks} (CLNs) \cite{pham2017column} which are deep architectures composed of several (feedforward) mini-columns, each of which represents an entity in the domain. Relationships between two entities are modeled through edges between mini-columns. 
The true power of column networks come from natural modeling of long-range inter-entity interactions with progressively deeper layers and have been successfully applied to collective classification tasks. However, CLNs rely on large amounts of data and incorporate little to no knowledge about the problem domain. While this may suffice for low-level applications such as image/video processing, it is a concern
in relational domains consisting of rich, semantic information.


Biasing the learners is necessary in order to allow them to inductively leap from training instances to true generalization over new instances ~\cite{Mitchell80}. 
While deep learning does incorporate one such bias in the form of domain knowledge (for example, through parameter tying or convolution, which exploits neighborhood information), we are motivated to develop systems that can incorporate richer and more general forms of domain knowledge. This is especially germane for deep relational models as they inherently construct and reason over richer representations. 

One way in which a human can guide learning is by providing {\em rules over training examples and features} \cite{shavlik89ebnn,towell1994knowledge,fung2003knowledge,kunapuli2010online}. 
Another way that has been studied extensively is expressing {\em preferences} within the preference-elicitation framework \cite{BoutilierEtAl06}. We are inspired by this form of advice as they have been successful within the context of inverse reinforcement learning \cite{KunapuliEtAl13}, imitation learning \cite{odomaaai15} and planning \cite{DasEtAl18}. 

The motivation for our approach is as follows: to develop a framework that {\bf allows a human to guide deep learning} by incorporating rules and constraints that define the domain and its aspects. Incorporation of prior knowledge into deep learning has begun to receive interest recently \cite{DingEtAl18}. 
However, in many such approaches, the guidance is not through a human, but rather through a pre-processing algorithm to generate guidance. Our framework is much more general, in that a domain expert provides guidance during learning. We exploit the rich representation power of relational methods to capture, represent and incorporate such rules into relational deep learning models.


We make the following contributions: (1) we propose the formalism of Knowledge-augmented Column Networks (K-CLN), (2) we present an approach to inject generalized domain knowledge in a CLN and develop the learning strategy that exploits this knowledge, and (3) we demonstrate, across two real problems, in some of which CLNs have been previously employed, the effectiveness and efficiency of injecting domain knowledge. Specifically, our results across the domains clearly show statistically superior performance with small amounts of data. As far as we are aware, this is the first work on human-guided CLNs. 

\begin{figure*}[h]
  \begin{minipage}[b]{0.6\textwidth}
    \centering
    \includegraphics[width=\columnwidth]{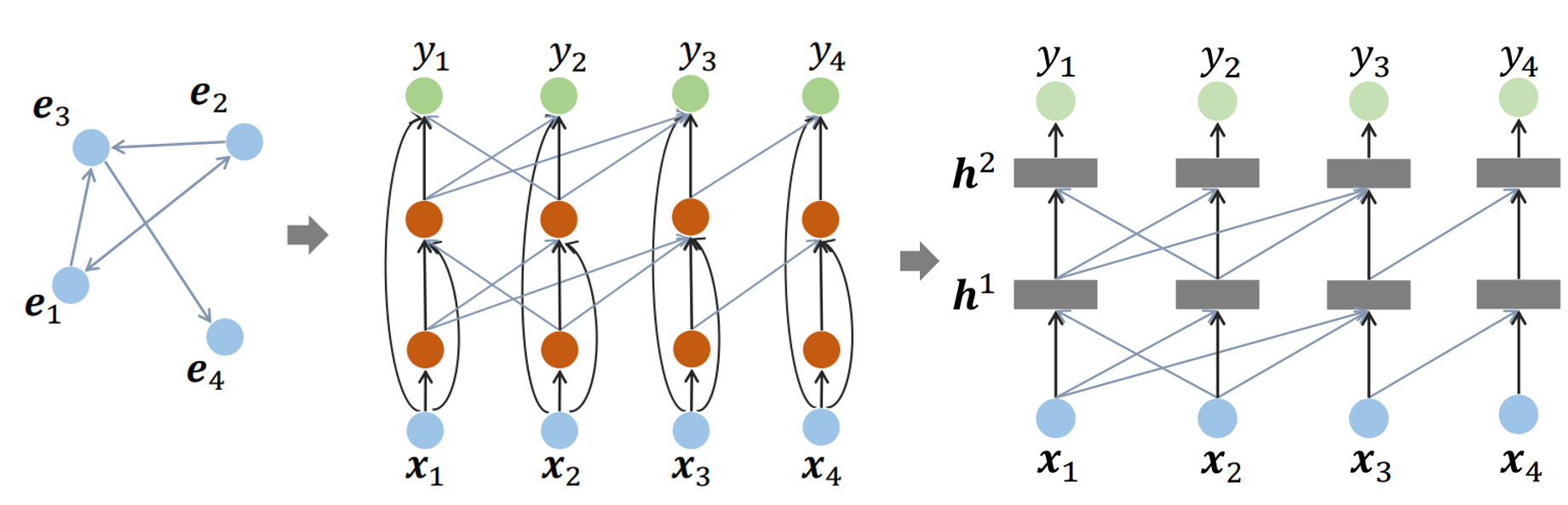}
    \captionof{figure}{Original Column network (diagram source: \cite{pham2017column})}
    \label{fig:CLN}
  \end{minipage}
  \begin{minipage}[b]{0.4\textwidth}
    \centering
    \includegraphics[width=0.9\columnwidth]{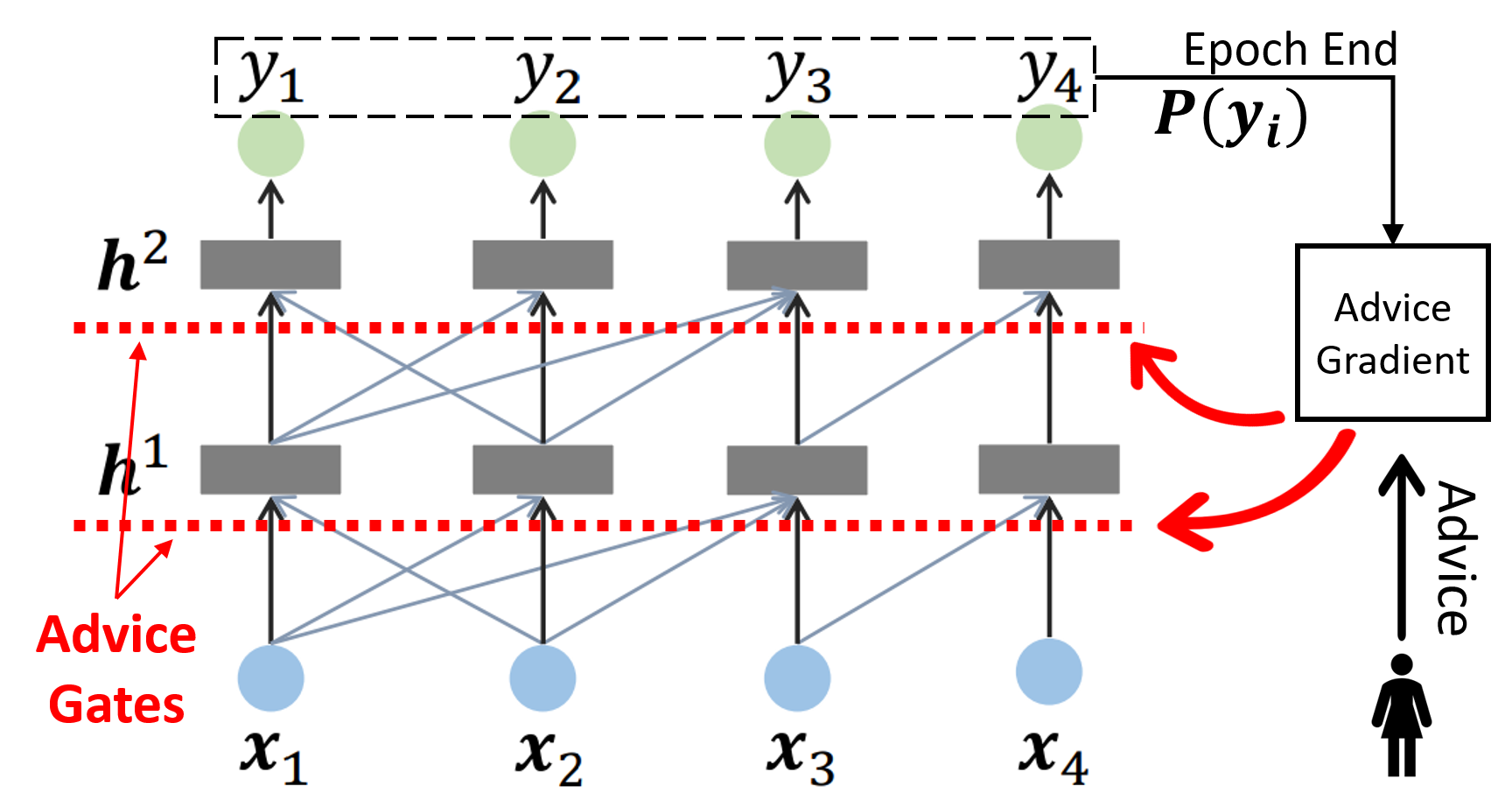}
      \captionof{figure}{K-CLN architecture}
      \label{fig:kcln}
    \end{minipage}
  \end{figure*}
\section{Knowledge-augmented Column Networks}
Column Networks~\cite{pham2017column} allow for encoding interactions/relations between entities as well as the attributes of such entities in a principled manner without explicit relational feature construction or vector embedding. This is important when dealing with structured domains, especially, in the case of collective classification. This enables us to seamlessly transform a multi-relational knowledge graph into a deep architecture making them one of the robust \textit{relational} deep models. Figure \ref{fig:CLN} illustrates an example column network, w.r.t. the knowledge graph on the left. Note how each entity forms its own column and relations are captured via the sparse inter-column connectors.

Consider a graph $\mathcal{G}=(V,A)$, where $V = \{e_i\}_{i=1}^{\left|V\right|}$ is the set of vertices/entities. 
$A$ is the set of arcs/edges between two entities $e_i$ and $e_j$ denoted as $r(e_i,e_j)$. Also, $\mathcal{G}$ is multi-relational, \textit{i.e.,} $r\in R$ where $R$ is the set of relation types in the domain. To obtain the equivalent Column Network $\mathcal{C}$ from $G$, let $x_i$ be the feature vector representing the attributes of an entity $e_i$ and $y_i$ its label predicted by the model\footnote{$\because i$ uniquely indexes $e_i$, we use $e_i$ and $i$ interchangeably}. $h_i^t$ denotes a hidden node w.r.t. entity $e_i$ at the hidden layer $t$ ($t=1, \ldots, T$  is the index of the hidden layers).
As mentioned earlier, the \textit{context} between 2 consecutive layers captures the dependency of the immediate neighborhood (based on arcs/edges/inter-column connectors). For entity $e_i$, the context w.r.t. $r$ and hidden nodes are computed as, 
\begin{align}
    \label{eqn:hiddencontext}
    c_{ir}^t = \frac{1}{\left|\mathcal{N}_r(i)\right|}\sum_{j\in \mathcal{N}_r(i)} h_j^{t-1}; \\
    h_i^t = g\left(b^t + W^t h_i^{t-1} + \frac{1}{z} \sum_{r\in R} V_r^t c_{ir}^t\right)
\end{align}
where $\mathcal{N}_r(i)$ are all the neighbors of $e_i$ w.r.t. $r$ in the knowledge graph $\mathcal{G}$. Note the absence of context connectors between $h^t_2$ and $h^t_4$ (Figure \ref{fig:CLN}, \textit{right}) since there does not exist any relation between $e_2$ and $e_4$ (Figure \ref{fig:CLN}, \textit{left}). 
The activation of the hidden nodes is computed as the sum of the bias, the weighted output of the previous hidden layer and the weighted contexts 
where $W^t \in \mathbb{R}^{K^t \times K^{t−1}}$ and $V^t_r \in R^{K^t\times K^{t−1}}$ are weight parameters and $b^t$ is a bias for some activation function $g$. $z$ is a pre-defined constant that controls the parameterized contexts from growing too large for complex relations. Setting $z$ to the average number of neighbors of an entity is a reasonable assumption.  
The final output layer is a softmax over the last hidden layer. 
\begin{equation}
    \label{eq:op} P(y_i = \ell|h_i^T) = softmax\left( b_l + W_l h_i^T \right)
\end{equation}
where $\ell\in L$ is the label ($L$ is the set of labels) and $T$ is the index of the last hidden layer. 

While CLNs are relation-aware deep models that can represent and learn from structured data faithfully, they are not devoid of limitations, especially the challenges of effective learning with sparse samples or systematic noise. Several approaches~\cite{jiang2018mentornet,goldberger2016training,miyato2018virtual} enable effective learning of deep models in presence of noise, however our problem setting is significantly different, due to  - 
    \textbf{{[(1) Type of noise]}}: we aim to handle systematic/targeted noise \cite{odom2018human}. It occurs frequently in real-world due to cognitive bias  or sample sparsity.
    \textbf{{[(2) Type of error]}}: Systematic noise leads to generalization errors.
    \textbf{{[(3) Structured data]}}: K-CLN works in the context of structured data (entities/relations). 
    Structured/relational data, thought crucial, is inherently sparse (most relations are false in the real world). 
\noindent \fbox{
\parbox{0.97\columnwidth}{
\noindent {\bf Given}: A sparse multi-relational graph $\mathcal{G}$, attributes $x_i$ of each entity (sparse or noisy) in $\mathcal{G}$, equivalent Column-Network $\mathcal{C}$ and access to a Human-expert\\
\noindent{\bf To Do:} More effective and efficient collective classification by knowledge augmented training of $\mathcal{C}(\theta)$, where $\theta = \langle\{W^t\}_1^T, \{V_r^t\}_{r\in R; t=1}^{t=T}, \{W_{\ell}\}_{\ell\in L}\rangle$ is the set of all the network parameters of the Column Network.
}}
To this effect we propose \textbf{K}nowledge-augmented \textbf{C}o\textbf{L}umn \textbf{N}etwork that incorporates human advice into deep models in a principled manner using a gated architecture, where \textit{`Advice Gates'} augment/modify the trained network parameters based on the advice.

\subsection{Knowledge Representation}
Any model specific encoding of domain knowledge, such as numeric constraints or modified loss functions etc., has several limitations, namely (1) counter-intuitive to the humans since they are domain experts and not experts in machine learning (2) the resulting framework is brittle and not generalizable.  Consequently, we employ preference rules (akin to IF-THEN statements) to capture human knowledge.
\begin{definition}
\label{def:pref}
A preference is a modified Horn clause,
\begin{align}
    \nonumber \mathtt{\land_{k, x} Attr_k(E_x) \land \ldots \land_{r\in R, x, y} r(E_x,E_y)} \Rightarrow \mathtt{[} & \mathtt{label(E_z,\ell_1)} \uparrow; \\ \nonumber &\mathtt{label(E_k,\ell_2) \downarrow]}
\end{align}
where $\ell_1,\ell_2 \in L$ and the $\mathtt{E_x}$ are variables over entities, $\mathtt{Attr_k(E_x)}$ are attributes of $E_x$ and $\mathtt{r}$ is a relation. $\mathtt{\uparrow}$ and $\mathtt{\downarrow}$ indicate the preferred non-preferred labels respectively. Quantification is implicitly $\forall$ and hence dropped. We denote a set of preference rules as $\mathfrak{P}$.
\end{definition}


\subsection{Knowledge Injection}
Given that knowledge is provided as \textit{partially-instantiated} preference rules $\mathfrak{P}$, more than one entity may satisfy a preference rule. Also, more than one preference rules may be applicable for a single entity.
The key idea is that we aim to consider the error of the trained model w.r.t. both the data and the advice. Consequently, in addition to the \textit{``data gradient"} as with original CLNs, there is a \textit{``advice gradient''}. This gradient acts a feedback to augment the learned weight parameters (both column and context weights) towards the direction of the \textit{advice gradient}. It must be mentioned that not all parameters will be augmented. Only the parameters w.r.t. the entities and relations (contexts) that satisfy $\mathfrak{P}$ should be affected. 
Let $\mathcal{P}$ be the set of entities and relations that satisfy the set of preference rules $\mathfrak{P}$. The hidden nodes (equation~\ref{eqn:hiddencontext}) can now be expressed as,
\begin{align}
  \label{eq:modhidden} \nonumber   h_i^t = g\left(b^t + W^t h_i^{t-1} \Gamma^{(W)}_i + \frac{1}{z} \sum_{r\in R} V_r^t c_{ir}^t \Gamma^{(c)}_{ir}\right)\\
  \text{s.t.}~ \Gamma_i, \Gamma_{i,r} = 
 \begin{cases}
                                   1 & \text{if $i,r \notin \mathcal{P}$} \\
                                \mathcal{F}(\alpha\nabla_i^{\mathfrak{P}}) & \text{if $i,r \in \mathcal{P}$}
  \end{cases}
\end{align}
where $i \in \mathcal{P}$ and $\Gamma^{(W)}_i$ and $\Gamma^{(c)}_{ir}$ are advice-based soft gates with respect to a hidden node and its context respectively. $\mathcal{F}()$ is some gating function, $\nabla_i^{\mathfrak{P}}$ is the \textit{``advice gradient''} and $\alpha$ is the trade-off parameter explained later. The key aspect of soft gates is that they attempt to enhance or decrease the contribution of particular edges in the column network aligned with the direction of the \textit{``advice gradient''}. We choose the gating function $\mathcal{F}()$ as an exponential $[\mathcal{F}(\alpha\nabla_i^{\mathfrak{P}}) = \exp{(\alpha\nabla_i^{\mathfrak{P}})}]$. The intuition is that soft gates are natural, as they are multiplicative and a positive gradient will result in $\exp{(\alpha\nabla_i^{\mathfrak{P}})} > 1$ increasing the value/contribution of the respective term, while a negative gradient results in $\exp{(\alpha\nabla_i^{\mathfrak{P}})} < 1$ pushing them down. We now present the \textit{``advice gradient''} (the gradient with respect to preferred labels). 

\begin{proposition}
\label{eq:grad}
Under the assumption that the loss function with respect to advice / preferred labels is a log-likelihood, of the form $\mathcal{L^\mathfrak{P}} = \log P(y_i^{(\mathfrak{P})}|h_i^T)$, then the advice gradient is,
$ \nabla_i^{\mathfrak{P}} = I({y_i^{(\mathfrak{P})}}) - P(y_i)$, 
where $y_i^{(\mathfrak{P})}$ is the preferred label of entity and $i\in \mathcal{P}$ and $I$ is an indicator function over the preferred label. For binary classification, the indicator is inconsequential but for multi-class scenarios it is essential ($I = 1$ for preferred label $\ell$ and $I=0$ for $L\setminus \ell$).
\end{proposition}

 An entity can satisfy multiple advice rules. So we take the most preferred label. 
 In case of conflicting advice (i.e. different labels are equally advised), we set the advice label to be the label given by the data, $y_i^{(\mathfrak{P})}=y_i$.


\begin{proposition}
\label{prop:balance}
Given that the loss function $\mathcal{H}_i$ of original CLN is cross-entropy (binary or sparse-categorical for the binary and multi-class prediction cases respectively) and the objective with respect to advice is log-likelihood, the functional gradient of the modified objective for K-CLN is, 
\begin{align}
\label{eqn:modgrad}
   \nonumber  \nabla(\mathcal{H}'_i) = & (1-\alpha)\left(y_iI - P(y_i|h^T)\right) + \alpha (I_i^{\mathfrak{P}}-P(y_i^{\mathfrak{P}}|h^T))\\
     = & (1-\alpha)\nabla_i + \alpha \nabla_i^{\mathfrak{P}}
\end{align}
where $0\leq\alpha\leq 1$ is the trade-off parameter between the effect of data and effect of advice, $I_i$ and $I_i^{\mathfrak{P}}$ are the indicator functions on the label w.r.t. the data and the advice respectively and $\nabla_i$ and $\nabla_i^{\mathfrak{P}}$ are the gradients, similarly, w.r.t. data and advice respectively.
\end{proposition}

Hence, it follows from Proposition~\ref{prop:balance} that the data and the advice balances the training of the K-CLN network parameters $\theta^\mathfrak{P}$ via the trade-off hyperparameter $\alpha$. When data is noisy (or sparse with negligible examples for a region of the parameter space) \textbf{the advice (if correct) induces a bias on the output distribution towards the correct label}. Even if the advice is incorrect, the network still tries to learn the correct distribution to some extent from the data (if not noisy). The contribution of the effect of data versus the effect of advice will primarily depend on $\alpha$. \textbf{If both the data and human advice are sub-optimal (noisy), the correct label distribution is not even learnable}. We exclude the formal proofs due to space limitation.

\setlength{\textfloatsep}{4pt}
\begin{algorithm}
\begin{algorithmic}[1]
\REQUIRE Knowledge graph $\mathcal{G}$, Column network $\mathcal{C}(\theta)$, Advice $\mathfrak{P}$, Trade-off $\alpha$
\STATE K-CLN $\mathcal{C}^{\mathfrak{P}}(\theta^{\mathfrak{P}}) \gets \mathcal{C}(\theta)$ 
\STATE Initialize parameters of K-CLN $\theta^{\mathfrak{P}} \gets \{0\}$ 
\STATE $\mathcal{M}^\mathcal{P} = \langle\mathcal{M}^W,\mathcal{M}^c,\mathcal{M}^{label}\rangle \gets$ \textsc{CreateMask}({$\mathcal{G},\mathfrak{P}$}) 
 \STATE Initial gradients @ epoch 0 $\forall i ~ \mathbf{\nabla}_{i,0}^{\mathfrak{P}} = 0$; $i \in \mathcal{P}$ 
\FOR{epochs k=1 to convergence} 
\STATE Get advice gradients $\nabla_{i,(k-1)}^{\mathfrak{P}}$ w.r.t. prev. epoch $k-1$
\STATE Gates $\Gamma^{\mathfrak{P}}_i, \Gamma^{\mathfrak{P}}_{i,r} \gets \exp{(\alpha \nabla_i^{\mathfrak{P}}\times \mathcal{M}_i^\mathcal{P})}$  
\STATE Train $\mathcal{C}^{\mathfrak{P}}$ using Equation~\ref{eq:modhidden}; Update $\theta^{\mathfrak{P}}$
\STATE Compute $\forall i ~ P(y_i)$ from $\mathcal{C}^{\mathfrak{P}}$ 
{for current epoch $k$}
\STATE Store $\forall i ~ \nabla_{i,k}^{\mathfrak{P}} \gets I({y_i^{(\mathfrak{P})}}) - P(y_i)$  
{using $\mathcal{M}^{label}$}
\ENDFOR
\STATE \textbf{return} {K-CLN $C^{\mathfrak{P}}$}
\end{algorithmic}
\caption{\underline{\textbf{K}}nowledge-augmented \underline{\textbf{C}}o\underline{\textbf{L}}umn \underline{\textbf{N}}etworks}
\label{algo:kcln}
\end{algorithm}

\subsection{The Algorithm}
Algorithm~\ref{algo:kcln} outlines the key steps involved in our approach. It trains a Column Network using both the data (the knowledge graph $\mathcal{G}$) and the human advice (set of preference rules $\mathfrak{P}$). It returns a K-CLN $\mathcal{C}^{\mathfrak{P}}$ where $\theta^{\mathfrak{P}}$ are the network parameters. As described earlier, the network parameters of K-CLN (same as CLN) are manipulated (stored and updated) via tensor algebra with appropriate indexing for entities and relations. Also recall that our gating functions are piece-wise/non-smooth and apply only to the subspace of entities, features and relations where the preference rules are satisfied. Thus, as a pre-processing step, we create tensor masks that compactly encode such a subspace with a call to the procedure \textsc{CreateMask()}, explained later. 
At the end of every epoch the output probabilities and the gradients are computed and stored in a shared data structure [\textbf{line: 10}] for computing advice gates in the next epoch. Rest of the training strategy is similar original CLN, except modified hidden units (Equation~\ref{eq:modhidden}) [\textbf{line: 8}] and data and advice trade-off parameter $\alpha$.

Procedure \textsc{CreateMask()} constructs the advice tensor mask(s) over the space of entities, features and relations/contexts, based on the advice rules, that are required to compute the gates. 
The main components are - 
    \textbf{(1)} The entity mask $\mathcal{M}^W$ ($\left|entities\right| \times \left|feature\right|$ tensor)  that indicates entity and feature indexes affected by the preferences;
    \textbf{(2)} The context mask $\mathcal{M}^c$ ($\left|entities\right| \times  \left|entities\right|$) which indicates the affected contexts/relations;
    \textbf{(3)} The label mask $\mathcal{M}^{label}$ stores the preferred label of the affected entities, in one-hot encoding.
Advice mask computation requires efficient satisfiability checking for each preference rule against the knowledge graph. We solve this via efficient subgraph matching proposed by Das et al.~\yrcite{DasAAAI19}. The masks are binary with $1$ encoding true and $0$ encoding false. 



\begin{figure*}[t]
\begin{minipage}{\textwidth}
    \centering
    \subfigure
    {
    \includegraphics[width=0.40\textwidth]{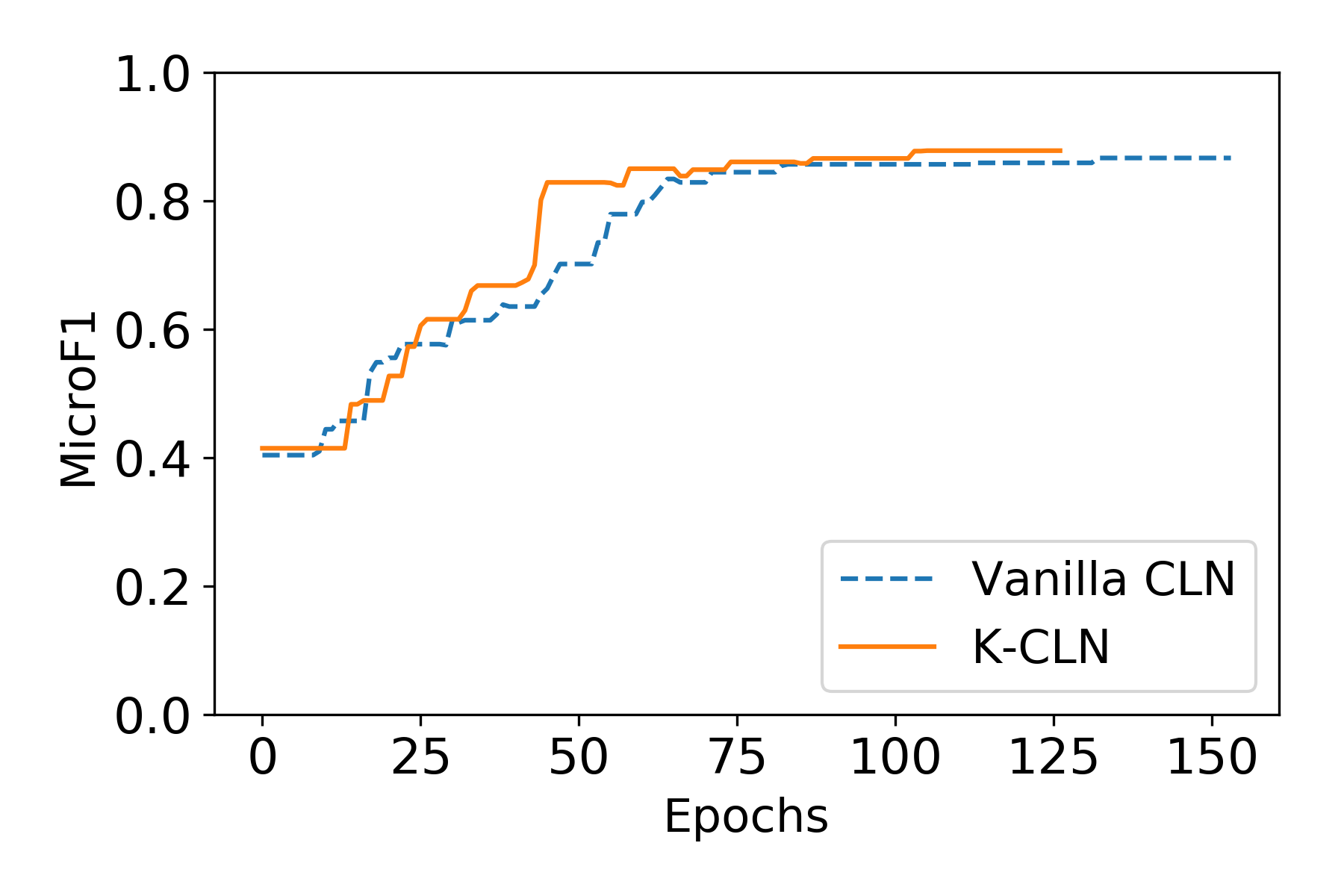}
    \label{fig:microPub}
    }
    \subfigure
    {
    \includegraphics[width=0.40\textwidth]{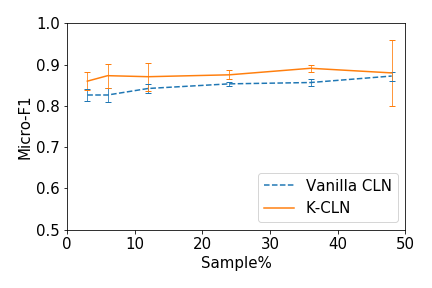}
    \label{fig:microPubSam}
    }
    \caption{\textbf{[Pubmed Diabetes publication prediction (multi-class)]} Learning curves - Micro-F1, \figleft ~ w.r.t. training epochs at 24\% (of total) sample, \figright ~ w.r.t. varying sample sizes [best viewed in color].}
    \label{fig:PubMed}
\end{minipage}
\begin{minipage}{\textwidth}
    \centering
    \subfigure
    {
    \includegraphics[width=0.40\textwidth]{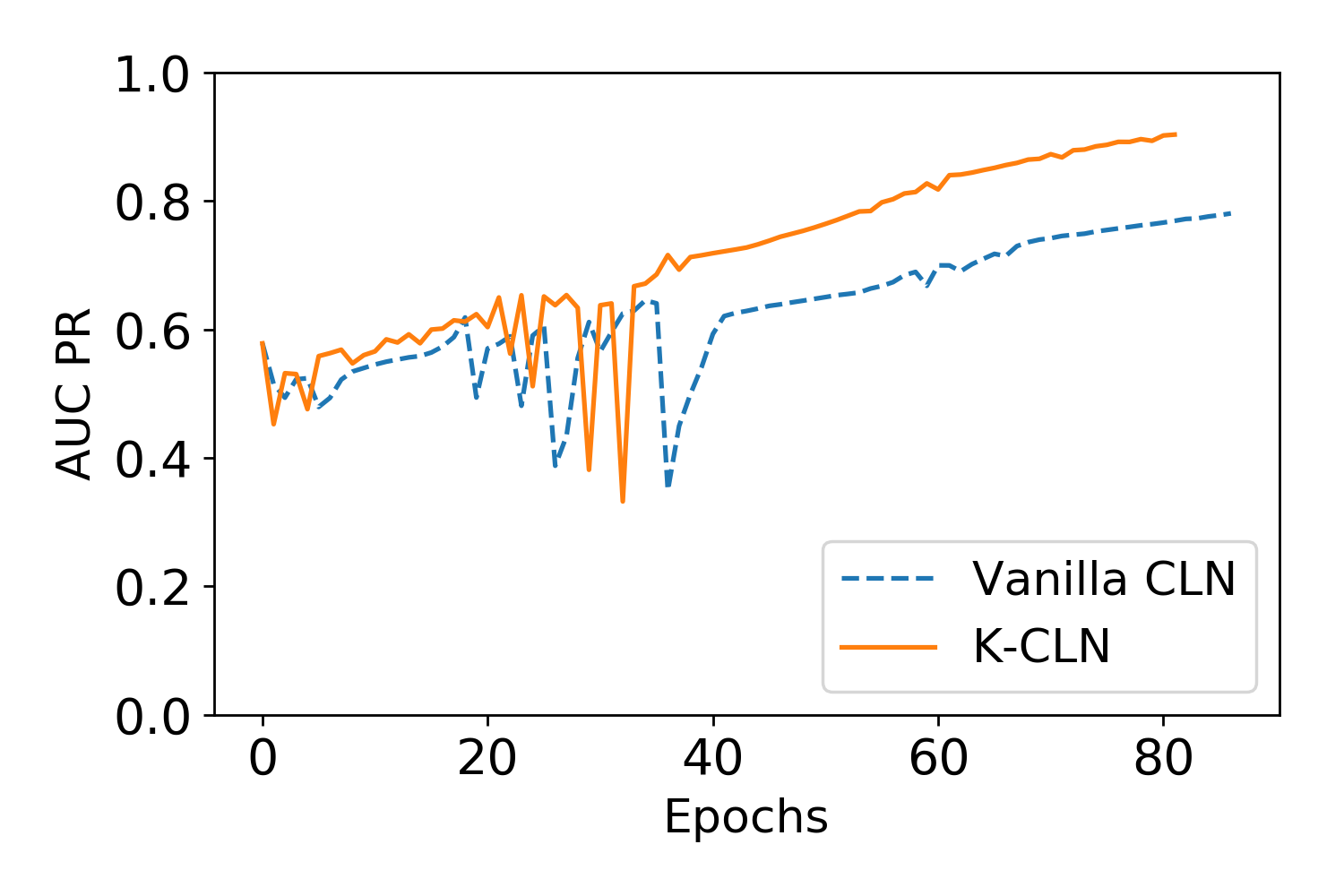}
    \label{fig:debateauc}}
    \subfigure
    {
    \centering
    \includegraphics[width=0.40\textwidth]{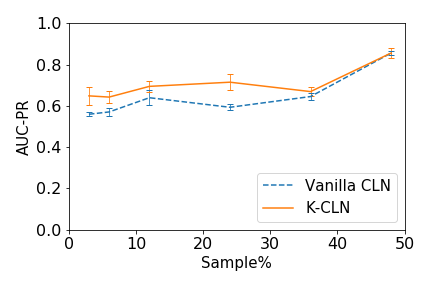}
    \label{fig:debateaucvary}}
    \caption{\textbf{[Internet Social debate stance prediction (binary class)]} Learning curves  - Micro-F1, \figtop ~ w.r.t. training epochs at 24\% (of total) sample, \figbottom ~ w.r.t. varying sample sizes [best viewed in color].}
    \label{fig:debate}
\end{minipage}
\end{figure*}

\begin{figure*}
\begin{minipage}{\textwidth}
\centering
\subfigure[F1 (varying sample \& $\alpha$)]{
\includegraphics[width = 0.40\columnwidth]{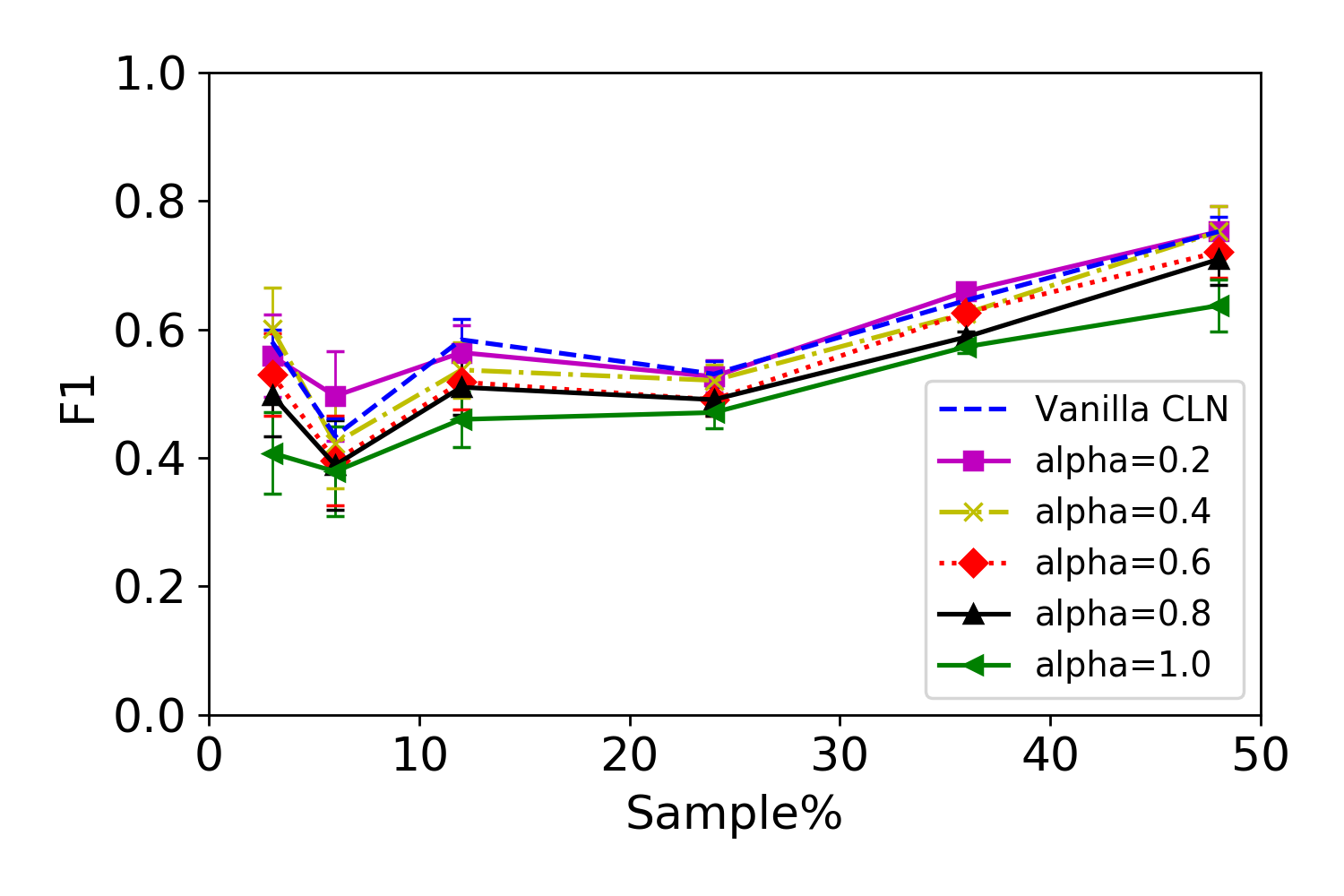}
}
\subfigure[AUC-PR (varying sample \& $\alpha$)]{
\includegraphics[width=0.40\columnwidth]{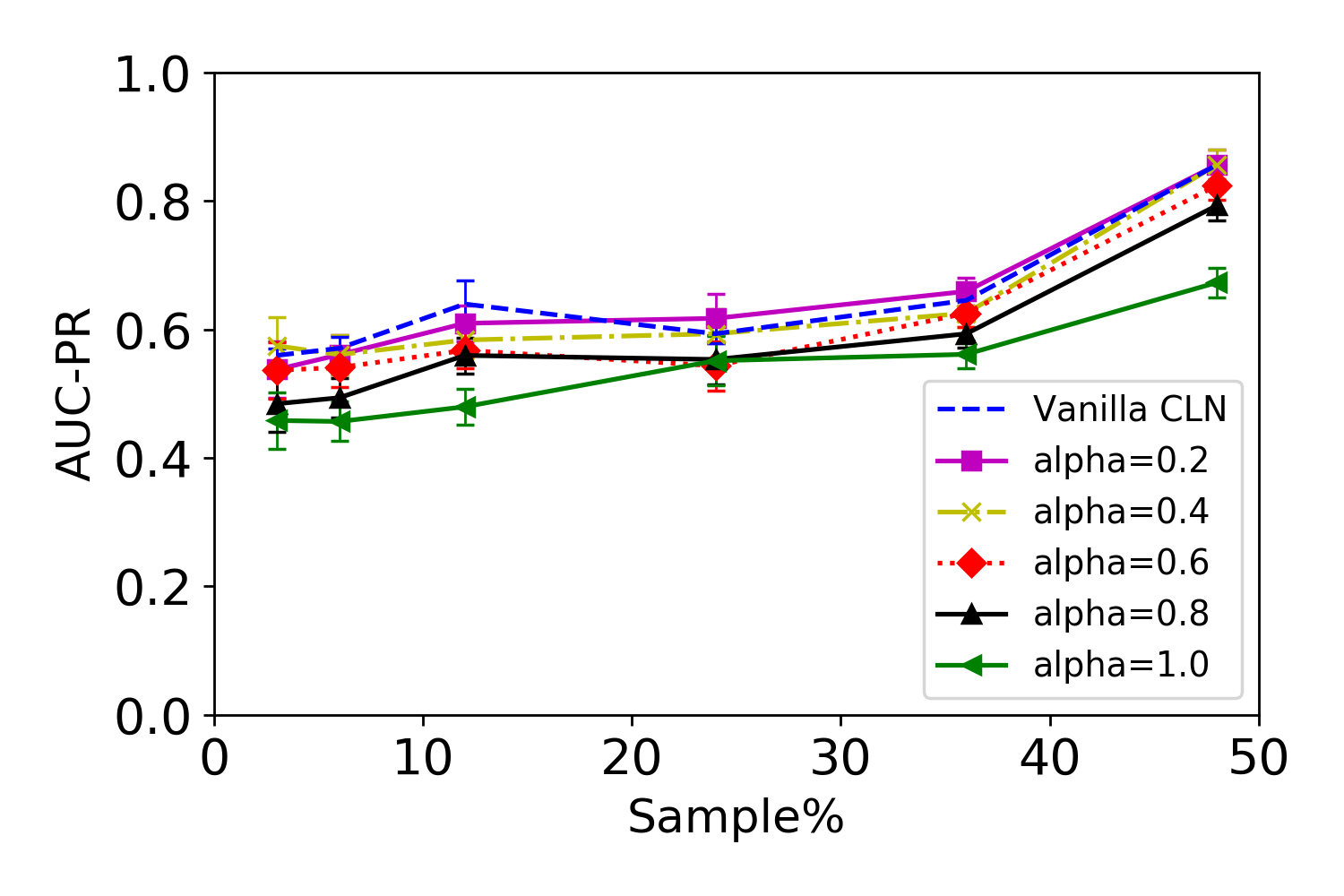}
}
\caption{Performance, F1 and AUC-PR, of K-CLN on \textbf{Internet Social Debates data set} across different sample sizes, with varying \textbf{\textit{trade-off parameter} $\alpha$} (on the advice gradient). Note that the advice here is incorrect/sub-optimal. $\alpha = 0$ has the same performance as no-advice (Vanilla CLN), hence not plotted.}
\label{fig:alphas}
\end{minipage}
\end{figure*}

\section{Experiments}
We investigate the following questions as part of our experiments, - 
    [\textbf{Q1}] Can K-CLNs learn effectively with noisy sparse samples i.e., performance?
    [\textbf{Q2}] Can K-CLNs learn efficiently with noisy sparse samples i.e., speed of learning?
    [\textbf{Q3}] How does quality of advice affect the performance of K-CLN i.e., reliance on robust advice?
We compare against the original Column Networks architecture with no advice\footnote{Vanilla CLN indicates original architecture \cite{pham2017column}} as a baseline. We show how advice/knowledge can guide model learning towards better predictive performance and efficiency, in the context of collective classification using Column Networks. 

\subsection{Experimental Setup}
\noindent \textbf{System:}
K-CLN has been developed by extending original CLN architecture, which uses \textit{Keras} as the functional deep learning API with a \textit{Theano} backend for tensor manipulation. We extend this system to include: (1) advice gradient feedback at the end of every epoch, (2) modified hidden layer computations and (3) a pre-processing wrapper to parse the advice/preference rules and create appropriate tensor masks. Since it is not straightforward to access final layer output probabilities from inside any hidden layer using keras, we use \textit{Callbacks} to write/update the predicted probabilities to a shared data structure at the end of every \textit{epoch}. Rest of the architecture follows from original CLNs. 
The \textit{advice masks} encode $\mathcal{P}$, \textit{i.e.}, the set of entities and contexts where the gates are applicable. 

\noindent\textbf{Domains:} We evaluate our approach on {\bf two relational} domains -- \textit{Pubmed Diabetes}, a multi-class classification problem and \textit{Internet Social Debates}, a binary classification problem. \textit{Pubmed Diabetes}\footnote{\url{https://linqs.soe.ucsc.edu/data}} is a citation network for predicting whether a peer-reviewed article is about \textit{Diabetes Type 1, Type 2 or none}, using textual features (TF-IDF vectors) from $19717$ pubmed abstracts. It comprises  articles, considered as an entities, with $500$ bag-of-words textual features (TF-IDF weighted word vectors), and $44,338$ citation relationships among each other. 
\textit{Internet Social Debates}\footnote{\url{http://nldslab.soe.ucsc.edu/iac/v2/}} is a data set for predicting stance (`for'/`against') about a debate topic from online posts on social debates. It contains $6662$ posts (entities) characterized by TF-IDF vectors, extracted from the text and header, and $\sim 25000$ relations of 2 types, {`sameAuthor'} and {`sameThread'}. 

\noindent\textbf{Metrics:} Following \cite{pham2017column}, we report micro-F1 score, \textit{which aggregates the contributions of all classes to compute the average F1 score}, for the multi-class problem and AUC-PR for the binary one. We use $10$ hidden layers and $40$ hidden units per column in each layer. All results are averaged over 5 runs and our settings are consistent with original CLN.

\noindent\textbf{Human Advice:} K-CLN can handle arbitrarily complex advice (encoded as preference rules). However, even with some relatively simple rules K-CLN is effective in sparse samples. \textit{For instance, in Pubmed, the longest preference rule used is, $\mathtt{HasWord(e_1,`fat')}$ $\land$ $\mathtt{HasWord(e_1,`obese')}$ $\land$ $\mathtt{Cites(e_2,e_1)}$ $\Rightarrow$ $\mathtt{label(e_2, type_2)}\uparrow$}. This simply indicates an article citing another one discussing obesity. is likely about Type2 diabetes, Expert knowledge from real physicians can thus, prove to be even more effective.  
Note that sub-optimal advice may lead to a wrong direction of the \textit{Advice Gradient}. However, since the data balances the effect of advice during training as shown by \cite{patrini2017making}, our soft gates do not alter the loss but instead promote/demote the contribution of nodes/contexts. 


\subsection{Experimental Results}
Our goal is to demonstrate the efficiency and effectiveness of K-CLNs with smaller set of training examples. Hence, we present the aforementioned metrics with varying sample size and with varying epochs and compare our model against \textit{Vanilla CLN}. We split the data sets into a training set and a hold-out test set with 60\%-40\% ratio. For varying epochs we only learn on 40\% of our training set (\textit{i.e.}, 24\% of the complete data) to train the model with varying epochs and test on the hold-out test set. Figures \ref{fig:PubMed} \figleft~ and \ref{fig:debate} \figleft~ illustrate the micro-F1 scores with \textit{varying epochs} for \textit{PubMed diabetes} and \textit{internet social debate} data sets resp. 
K-CLN converges {\bf significantly faster} (less epochs), at times, with better predictive performance at convergence 
which shows that K-CLNs learn more \textit{efficiently} with noisy sparse samples thereby answering \textbf{(Q1)} affirmatively. 

Effectiveness of K-CLN is illustrated by its performance with respect to the varying sample sizes of the training set, especially in the low sample ranges. The intuition is, \textit{domain knowledge should help guide the model to learn better when the amount of training data available is small}. K-CLN is trained on gradually varying sample size from 5\% of the training data (3\% of the complete data) till 80\% of the training data (48\% of complete data) and tested on the hold-out test set. Figures \ref{fig:PubMed} \figright~ and \ref{fig:debate} \figright~ present the micro-F1 with varying sample sizes for \textit{PubMed diabetes} and \textit{internet social debate} respectively.
For \textit{internet social debate stance prediction}, K-CLN outperforms Vanilla CLN with all sample sizes lower than, approximately, $35\%$. However, in case of \textit{PubMed}, K-CLN outperforms Vanilla CLN for all sample sizes we experimented with, thus answering \textbf{(Q2)} affirmatively. K-CLNs learn \textit{effectively} with noisy sparse samples.

An obvious question that will arise is -- {\em how robust is our learning system to that of noisy/incorrect advice?} Conversely, {\em how does the choice of $\alpha$ affect the quality of the learned model?}
To answer these questions specifically, we performed an additional experiment on the \textbf{Internet Social Debates} domain by augmenting the learner with incorrect advice. This incorrect advice is essentially created by changing the preferred label of the advice rules to incorrect values (based on our understating). Also, recall that  the contribution of advice is dependent on the trade-off parameter $\alpha$, which controls the robustness of K-CLN to advice quality. Consequently, we experimented with different values of $\alpha$ ($0.2,0.4,\ldots,1.0$), across varying sample sizes.

Figure~\ref{fig:alphas} shows how with higher $\alpha$ values the performance deteriorates due to the effect of noisy advice. $\alpha=0$ is not plotted since the performance is same as no-advice/Vanilla CLN. Note that with reasonably low values of $\alpha = 0.2, 0.4$, the performance does not deteriorate much and is, in fact, better in some samples. Thus with reasonably low values of $\alpha$ K-CLN is robust to quality of advice \textbf{(Q3)}. We picked one domain to present the results of this robustness but have observed similar behavior in both the domains. These experiments empirically support our theoretical analysis (Proposition~\ref{prop:balance}). We found that when $\alpha \leq 0.5$, K-CLN performs well even with noisy advice. In the earlier experiments where we use potentially good advice, we report the results with $\alpha=1$, So it is reasonable to assign higher weight to the advice and the contribution of the entities and relations/contexts affected by it, given the advice is noise-free. Also, note that the drop in performance towards very low sample sizes (in Figure~\ref{fig:alphas}) highlights how learning is challenging in the noisy-data and noisy-advice scenario. This aligns with our general understanding of most human-in-the-loop/advice-based approaches in AI. Trade-off between data and advice via a weighted combination of both is a well studied solution in related literature \cite{OdomNatarajan18} and, hence, we adapt the same in our context. Tracking the expertise of humans to infer advice quality is an interesting future research direction.

\section{Conclusion}
We considered the problem of providing guidance for CLNs. Specifically, inspired by treating the domain experts as true domain experts and not CLN experts, we developed a formulation based on {\em preferences}. This formulation allowed for natural specification of guidance. We derived the gradients based on advice and outlined the integration with the original CLN formulation. Our initial evaluation across a couple of domains clearly demonstrate the effectiveness and efficiency of the approach, specifically in knowledge-rich, data-scarce problems. We are also experimenting on a few more domains and the results will be included in the full version of the paper. Exploring other types of advice including feature importance, qualitative constraints, privileged information, etc. is a potential future direction. Scaling our approach to web-scale data is a natural extension. Finally, extending the idea to other deep models and applications to more real domains remains an interesting direction for future research.

\paragraph{Acknowledgements: }
MD, GK \& SN gratefully acknowledge the support of CwC Program Contract W911NF-15-1-0461 with the US Defense Advanced Research Projects Agency (DARPA) and the Army Research Office (ARO). SN also acknowledges the NSF grant IIS-1836565 and AFOSR award FA9550-18-1-0462. DSD acknowledges the National Institute of Health (NIH) grant no. R01 GM097628. Any opinions, findings and conclusion or recommendations are those of the authors and do not necessarily reflect the view of the DARPA, ARO, AFOSR or the US government.

\bibliographystyle{icml2019}
\bibliography{biblio}

\end{document}